\documentclass[a4paper]{article}
\usepackage{pmstyleeng}
\usepackage{cite}
\usepackage{indentfirst}
\usepackage{booktabs} 
\usepackage{makecell}
\usepackage{indentfirst}
\usepackage{algorithm}
\usepackage{algorithmicx}
\usepackage{algpseudocode}
\usepackage{amsmath}
\usepackage{CJK}
\usepackage{graphicx} 
\usepackage{epstopdf}
\usepackage{amsmath}
\usepackage{setspace}

\begin{document}
\hyphenation{SPbU}

%
\udk{}


\author{Zhang~Y., Li~Y.*, Li~Y., Guo~Z., Zhang~D.}

\title{A Review of Adversarial Attacks in Computer Vision}
\maketitle
\begin{spacing}{0.8}
\renewcommand{\thefootnote}{ }
{\footnotetext{{\it Zhang Yutong} -- graduate student, Harbin Institute of Technology; e-mail: 22s112078@stu.hit.edu.cn}}
{\footnotetext{{\it Li Yao} -- assistant professor, Harbin Institute of Technology; e-mail: yaoli0508@hit.edu.cn, corresponding author}}
{\footnotetext{{\it Li Yin} -- associate professor, Harbin Institute of Technology; e-mail: dr.liyin@hit.edu.cn}}
{\footnotetext{{\it Guo Zhichang} -- professor, Harbin Institute of Technology; mathgzc@hit.edu.cn}}
{\footnotetext{{\it Zhang Dazhi} -- professor, Harbin Institute of Technology; zhangdazhi@hit.edu.cn}}
\end{spacing}





\begin{spacing}{0.8}
\section{Introduction}
\end{spacing}
Deep neural networks have been widely used in various downstream tasks, especially those safety-critical scenario such as autonomous driving, but deep networks are often threatened by adversarial samples \cite{szegedy2013intriguing}. Such adversarial attacks can be invisible to human eyes, but can lead to DNN misclassification, and often exhibits transferability between deep learning and machine learning models \cite{goodfellow2014explaining} and real-world achievability \cite{kurakin2018adversarial}.

Adversarial attacks can be divided into \textbf{white-box attacks} (Section \ref{white-box-attack}), for which the attacker knows the parameters and gradient of the model, and \textbf{black-box attacks} (Section \ref{black-box-attack}), for the latter, the attacker can only obtain the input and output of the model. In terms of the attacker's purpose, it can be divided into targeted attacks and non-targeted attacks, which means that the attacker wants the model to misclassify the original sample into the specified class, which is more practical, while the non-targeted attack just needs to make the model misclassify the sample. The black box setting is a scenario we will encounter in practice.

Black-box attacks can also be divided into \textbf{query-based attacks}, which require a lot of repeated query model output to adjust perturbations, while \textbf{transfer-based} attacks do not, which makes the latter easier to do because too many queries are not allowed in practice.Transfer-based attacks often require the use of a white-box surrogate model to create adversarial perturbations, which are mostly developed from existing white-box attacks.

In terms of the way adversarial perturbations are generated, there are \textbf{optimization-based methods} (Section \ref{transfer-opt-start} - Section \ref{transfer-opt-end}) and \textbf{generative methods} (Section \ref{transfer-gen}). Optimization-based methods uses model gradients to iteratively change the specified sample to obtain perturbation, which is the most common way. In order to improve the transferability of adversarial samples, subsequent work combined optimization-based adversarial attacks with various means, such as adding momentum during iteration, performing input transformations, replacing losses, or building auxiliary classifiers. Compared to generative methods, they are still slightly inferior in target attack scenarios, and they need to iteratively create perturbations for each specified sample, while generative methods can be generalized on more samples, which could help to produce \textbf{Universal Adversarial Pertubations} (UAPs) (Section \ref{uap}). UAP aims to solve the problem that traditional optimized adversarial attacks are Instance-dependent, and it is hoped that the generated adversarial perturbation is effective for as many samples as possible.

For different tasks, adversarial attacks on traditional image classification tasks, adversarial attacks in \textbf{object detection} (Section \ref{obj-dec}) (including images and videos) and adversarial attacks in \textbf{semantic segmentation} (Section \ref{seg}) have all been studied. Object detection models tend to be more vulnerable because they need to predict the class and location, which is equivalent to both classification and regression tasks, while image classification models only need to predict classes. Semantic segmentation is a regression problem, which also makes it more vulnerable to adversarial attacks. Because regression problems are not like classification problems, the prediction result of classification problems does not change continuously with the change of input, but has a threshold, but in regression problems, the prediction result will change continuously as the input changes, which makes the model in the regression problem less robust and less resistant to adversarial attacks.

\section{Adversarial Attack Methods}
\subsection{White-box attacks}\label{white-box-attack}
White-box attack is the most common way to generate adversarial samples, which requires full access to the target model under attack, that is, knowing the structure of the model and the specific parameters of each layer. The attacker modifies the input data at the bit level according to the structure and parameters of the target model so that the data is misjudged by the target model. Mostly, white-box attacks generate adversarial perturbations for each single input image, but there are also general adversarial perturbations for target models and entire datasets. The success rate of white-box attacks can be very high since all the details of the target model is known. Once an adversarial perturbation is found, a better attack can be obtained by continuously reducing the disturbance. 
\begin{spacing}{0.95}
There are two main mainstream white-box adversarial attack algorithms: 1) optimization-based attack algorithms and 2) gradient-based attack algorithms.

\subsubsection{Optimization-based white-box attacks} 
\textbf{(1) Box-constrained L-BFGS}
\\
\end{spacing}
In 2013, Szegedy et al. \cite{szegedy2013intriguing} showed for the first time that a neural network can be misled to make a misclassification by adding a small amount of perturbation to an image that is not perceptible to humans. They first tried to solve the equation for the minimum perturbation that would allow the neural network to make a misclassification. Because the complexity of the problem was too high, they turned it to a simplified convex version, i.e. finding the minimum loss function additive term that allows the neural network to make a misclassification.

\begin{align}
\text{Minimize}_r& \quad c||r||_\infty+ loss_f(x+r,y^t) \\
\text{s.t.}&\quad x+r\in[0, 1]^m \notag
\end{align}

In the above function, $loss_f(x+r, y^t)$ is the cross-entropy loss that misled the model classifying adversarial sample to the target class $y^t$. The adversarial sample is constrained at $[0,1]$ to stay in the intensity range of images. 
The optimization problem is solved by first fixing the hyperparameter $c$ to find the optimal solution for the current $c$, and then finding the optimal adversarial perturbation $r$ that satisfies the condition by performing a linear search on $c$ to obtain the final adversarial sample $x+r$.
\\

\textbf{(2) C\&W}
\\

Carlini and Wagner \cite{carlini2017towards} propose the C\&W attack algorithm, which uses $L_0$, $L_2$, $L_\infty$ norms to generate adversarial samples with restrictions on perturbations respectively. It is one of the most powerful target attack algorithms available, based on an improved version of the Box-constrained L-BFGS algorithm. It considers the optimization problem to find an adversarial sample for a image $x$ as
\begin{align}
\text{minimize}_\delta& \quad D(x,x+\delta)\\
\text{s.t.}&\quad C(x+\delta)=t \notag\\
&\quad x+\delta\in[0,1]^n \notag
\end{align}
where the objective is to find a $\delta$ such that $D(x,x+\delta)$ is minimal and $D$ is some distance indicator that can be $L_0$, $L_2$, $L_\infty$ norm. The above objective function is difficult to optimize because the restrictions are highly non-linear. Thus, the first constraint is expressed in a different form as an objective function $f$ such that when $C(x + \delta) = t$ is satisfied,  $f(x + \delta)\leq 0$ is also satisfied, where $f$ has a wide range of choices. In the original paper, the empirical best $f$  is given by
\begin{align}
f(x^A)=\max(\max\{Z(x^A)_i:i\ne t \}-Z(x^A)_{t}, -k)
\end{align}
where $Z(x^A)=Logits(x^A)$ denotes the output of the previous layer of Softmax, $i$ denotes the label category, and k denotes the attack success rate of the adversarial sample, with larger generated adversarial perturbation having a higher attack success rate. 
\subsubsection{Gradient-based white-box attacks} 
\textbf{(1) FGSM}
\\

GoodFellow et al.\cite{goodfellow2014explaining} developed a method that efficiently computes the adversarial perturbation and referred as Fast Gradient Method (FGSM). In FGSM the amount of change in the adversarial perturbation is aligned with the direction of change in the gradient of the model loss. In an un-targeted attack, the gradient of the model loss function about the input $x$ is made to vary in the upward direction of the perturbation to achieve the effect of allowing the model to misclassify. The perturbations generated are adversarial perturbations subject to the $L_\infty$ norm constraint
\begin{align}
x^A=x+\alpha \; sign(\nabla _xJ(\theta; x,y))
\end{align}
where $sign(\cdot)$ is the sign function and $\alpha$ is the hyperparameter, denoted as the step size of the one-step gradient. 

FGSM generates adversarial samples with good migration attack capabilities, and its work has far-reaching influence. Most of the attack algorithms based on gradient optimization that have appeared since then are variants of the FGSM algorithm. However, since the gradient is calculated only once, its attack capability is limited. The premise of the successful application of this method is that the gradient direction of the loss function is linear in the local interval. In the nonlinear optimization interval, the adversarial examples generated by large stepwise optimization along the direction of gradient change cannot guarantee the success of the attack.
\\

\textbf{(2) I-FGSM}
\\

I-FGSM is an iterative FGSM algorithm that makes the linearity assumption hold approximately by making the optimization interval smaller. The objective-free attack of the I-FGSM algorithm is described as
\begin{align}
x^{A}_0 &= x \notag \\
x^{A}_{N+1} &= clip_{x,\varepsilon} \{x^{A}_N+\alpha \; sign(\nabla_x J(\theta; x^{A}_N,y_{true}))\}
\end{align}
where the step size is denoted by $\alpha$ and the number of iterations is denoted by $N$. As the perturbation amplitude can be assigned to each iteration, the step size $\alpha$ and $N$ can be set with $\alpha = e_N$ given $e$. The $clip{\cdot}$ function replaces the perturbation which larger than $\varepsilon$ with $\varepsilon$, because as the number of iterations increases, some of the pixel values may overflow. 

Kurakin et al. \cite{kurakin2018adversarial} build on the I-FGSM algorithm by targeting the lowest confidence category. The target category $y$ of the attack was specified as the category label $y_{LL}$ that the original sample outputs with the lowest confidence on the model. This targeting approach generates adversarial samples that misclassify the model to classes that are significantly different from the correct class, and the effect of the attack is more destructive. During the target attack, the direction of change of the perturbation is aligned with the direction of gradient descent of the model loss function with respect to the input, and the objective of its optimisation takes this form:
\begin{align}
x^{A}_0&=x \notag \\
x^{A}_{N+1}&=clip_{x,\varepsilon} \{x^{A}_N-\alpha \; sign(\nabla_x J(\theta; x^{A}_N,y_{LL}))\} 
\end{align}
The above method is a prototype of the targeted attack. Many of the current gradient-based targeted attack algorithms basically generate the attack image by replacing the real label with the target label and calculating the gradient.
\\

\textbf{(3) PGD}
\\

Projected Gradient Descent (PGD)\cite{madry2017towards} is an iterative attack that usually works better than FGSM and is one of the benchmark testing algorithms for assessing the robustness of a model. The PGD algorithm avoids encountering saddle points by doing multiple iterations, taking one small step at a time and randomly initializing the noise on the $x^A$ obtained in the previous iteration at each iteration. The perturbation is clipped to a prescribed range. Because the derivative of loss with respect to the input is fixed if the target model is linear, the direction of decline in loss is clear and the direction of perturbation does not change with the iterations, whereas for a non-linear model, the direction may not be exactly correct by doing just one iteration.

\subsubsection{Other white-box attack methods} 
\textbf{(1) DeepFool}
\\
\begin{spacing}{0.95}
Moosavi-Dezfooli et al.\cite{moosavi2016deepfool} generate minimum normative adversarial perturbations by an iterative computational method that gradually pushes images that lie within the classification boundary outside the boundary until a misclassification occurs. The authors show that the perturbations generated by Deepfool  are smaller than FGSM, while having similar deception rates.
For example, given a classifier $\hat{k}(x)=sign(f(x))$ in a binary classification problem $f(x)=\omega x+b$, the decision boundary of the classifier is the separating affine hyperplane $F=\{x:f(x)=0\}$. The minimum perturbation to change the classification of $x$ is to move $x$ onto the decision boundary, i.e. the minimum cost is the orthogonal projection $r_*(x_0)$ on $F$ for $x_0$:
\begin{align}
  &r_*(x_0) \equiv \arg\min\limits_{r} ||r||_2 \\
  &\text{s.t.} \quad sign(f(x_0+r)) \ne sign(f(x_0)) \notag
\end{align}

When solving a binary classification problem with non-linear decision boundaries, the minimum perturbation $r_*(x_0)$ of $x_0$ is approximated by the iterative process. The model is considered approximately linear during each iteration. The minimum distance of the data point $x_0$ corresponding to this iteration after the perturbation at this point is:

\begin{align}
&r_i(x_i)=\arg\min\limits_{r} ||r_i||_2  \\
&\text{s.t.} \quad f(x_i)+\nabla f(x_i)^Tr_i=0 \notag
\end{align}

The resulting closed-form solution yields the minimum perturbation in the set $r_i(x_i)=-\frac{f(x_i)}{||\nabla f(x_i)||^2_2}\nabla f(x_i)$ for the nonlinear decision boundary, and the perturbation $r_i$ obtained at each iteration is accumulated to obtain the minimum perturbation for the current point $x_0$.
\\

\textbf{(2) UAPs}
\\

Moosavi-Dezfooli et al. \cite{moosavi2017universal} found the existence of generic adversarial perturbations in deep learning models that are independent of the input samples, which are related to the target model structure and dataset features. Perturbationsthat enable attacks on any image can be generated that are also virtually invisible to humans. The approach used is similar to DeepFool in that it uses adversarial perturbations to push images out of classification boundaries, although the Universal Adversarial Perturbations algorithm (UAPs)  generates generic adversarial perturbations by iteratively computing over a small number of sampled data points, with the same perturbation targeting all images, and this perturbation approach has been shown to generalise to other networks.
\\

\textbf{(3) ATNs}
\\

Baluja and Fischer train multiple forward neural networks to generate adversarial samples that can be used to attack one or more networks \cite{baluja2017adversarial}. The Adversarial Transformation Networks algorithm (ATNs) generates adversarial samples by minimizing a joint loss function that has two components, the first of which keeps the adversarial samples similar to the original image, and the second of which causes the adversarial samples to be misclassified. ATNs can be targeted or untargeted and also trained in either a white-box or black-box manner.
\\

\textbf{(4) JSMA}
\\

A common approach used in the adversarial attack literature is to restrict the value of the $L_\infty$ or $L_2$ norm of the perturbation so that the perturbation in the adversarial sample is not perceptible. However, Jacobian-based Saliency Map Attack (JSMA) \cite{papernot2016limitations} proposes a method that restricts the $L_0$ norm by changing the value of only a few pixels, rather than perturbing the whole image. 

This is effectively a greedy algorithm that uses a saliency map to represent the degree of influence of the input features on the predicted outcome, modifying one pixel of a clean image at a time, and then calculating the bias derivative of the output of the final layer of the model for each feature of the input. With the resulting forward derivatives, the significance map is calculated. Finally, the significant graph is used to find the input features that have the greatest influence on the model's output, and by modifying these features that have a greater influence on the output, a valid counter sample is obtained.

\subsection{Black-box attacks}\label{black-box-attack}

Before the advent of black-box attacks, the generation of adversarial samples was based on a white-box approach, i.e. the attacker was fully aware of the structure of the model and the parameters such as the weights. In practice, however, this ideal condition is almost non-existent and it is almost impossible for the attacker to obtain detailed information about the model.

\subsubsection{Nicolas Papernot's attack}
\begin{figure}[h]
    \caption{Training of the substitute DNN $F$}
    \centering
    \includegraphics[width=\textwidth]{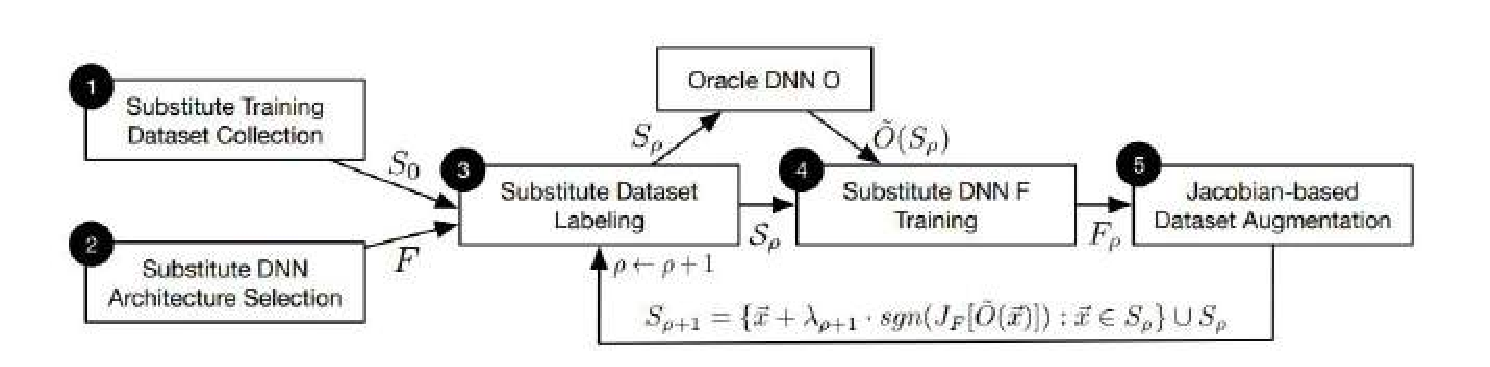}
    \label{img1}
\end{figure}

\end{spacing}
Nicolas Papernot et al propose a black-box attack based on training an alternative model that performs the same task as the target model that one wants to attack, generating adversarial samples based on the current model, which are eventually used to attack the original target model \cite{papernot2017practical}. However, training an alternative model $F$ to approximate the original model oracle $O$ is challenging because (a) the structure needs to be chosen for the alternative model without knowing the structural information of the original model. \\
(b) the number of queries (inputs and outputs) to the original model needs to be limited in order to ensure that this approach is feasible and easy to handle.
\begin{spacing}{0.95}
They overcome these challenges primarily by introducing a synthetic data generation technique, known as Jacobian-based dataset augmentation. This technique is not designed to maximize the accuracy of the alternative model, but to allow the alternative model to approximate the decision boundaries of the original model. The main training process is shown in Figure \ref{img1}.

\vspace*{1cm}

Initial Collection : The attacker collects a very small set $S_0$ representing the input domain. For example, if the original model is used to classify handwritten numbers, the attacker collects 10 images for each number from 0-9 and the distribution of this set does not have to match the distribution of the target model's training set.

Architecture Selection: Selects an architecture for the alternative model $F$. It should be noted that the type, number of layers and size of the model have little influence on the success of the attack.
\end{spacing}
Substitute Training: Iteratively train the substitution model $ F_{\rho}$ to improve its accuracy. $\rho$ is the number of iterations.

Labeling: The initial set of alternative inputs $S_{\rho}$ is fed into the model $O$ and the output is used as a label for the sample.

Training: Training the model $F_{\rho}$ with the labeled alternative training set.

Augmentation: The current dataset $S_{\rho}$ is expanded using the dataset expansion technique mentioned previously to obtain $S_{\rho+1}$, a new set that better represents the decision boundary of the model. Repeat the above Labeling and Training process with the set $S_{\rho+1}$.

The above process was repeated several times to improve the accuracy of $F$ and to make its decision boundary more similar to model $O$.
\begin{spacing}{0.95}
Nicolas Papernot use the trained alternative model $F$ to generate adversarial samples and mention two generation strategies:
The algorithm of Goodfellow et al:
\begin{equation}
    \delta_{\vec{x}} = \varepsilon sgn (\nabla_{\vec{x}} c(F,\vec{x},y))
\end{equation}
The algorithm of Papernot et al:
\begin{equation} 
S(\vec{x},t)[i]=
    \begin{cases}
         0 \  {\rm if} \  \frac{\partial F_t}{\partial \vec{x}_i}(\vec{x}) < 0 \ {\rm or} \sum_{j\neq t} \frac{\partial F_j}{\partial \vec{x}_i}(\vec{x}) > 0 \\ \frac{\partial F_t}{\partial \vec{x}_i}(\vec{x})|\sum_{j\neq t} \frac{\partial F_j}{\partial \vec{x}_i}(\vec{x})| \rm  \ otherwise   
    \end{cases}
\end{equation}

Each algorithm has its advantages and disadvantages. the Goodfellow algorithm is well suited to quickly producing many adversarial samples with relatively large perturbations and is therefore potentially easier to detect. the Papernot algorithm reduces perturbations at the expense of greater computational cost.

They validated the attack design by targeting a remote DNN provided by MetaMind, forcing it to misclassify 84.24\% of the adversarial samples. They also extensively calibrated our algorithm and generalised it to other ML models by instantiating it against Amazon and Google-hosted classifiers with success rates of 96.19\% and 88.94\%.

\subsubsection{NES}

Threat models for real-world systems are usually more restrictive than typical black-box models, such as limiting the number of queries, and when accessing a model excessively frequently, the model may self-lock to disallow access \cite{ilyas2018blackbox}. Andrew Ilyas et al propose three realistic threat models:

Query-limited setting: The attacker has a limited number of query results for the classifier. The limitation on the number of queries may be the result of restrictions on other resources, e.g., time limits and money limits.
\end{spacing}
Partial-information setting: local models may output probabilities for all categories, but for APIs may output only the top k categories, i.e. a score that does not sum to 1 in the output category, to indicate relative confidence in the prediction;

Label-only setting: In a label-only setting, an attacker cannot access category probabilities or scores. Instead, only a list of k inferred labels ordered by their predicted probabilities can be accessed.

In response to these three problems, Andrew Ilyas et al develop new attacks that fool the classifier under these more restrictive threat models, suggesting that previous approaches are impractical or ineffective.

In the query restriction setting, the attacker has a query budget $L$ that aims to cause targeted misclassification in $L$ or fewer queries. To attack this setting, Andrew Ilyas use "standard" first-order techniques to generate adversarial samples, replacing the gradient of the loss function with a gradient estimate, which is approximated by the query classifier rather than computed by automatic differentiation. Andrew Ilyas detail an algorithm for efficiently estimating gradients from queries based on the Natural Evolutionary Strategies approach of Wierstra et al. and then show how the estimated gradients can be used to generate adversarial samples.
\\
\begin{spacing}{0.95}
\textbf{Query-limited setting}
\\

To estimate the gradient, Andrew Ilyas use NES, a derivative-free optimization method based on the idea of a search distribution $\pi(\theta|x)$. Instead of maximizing the objective function $F(x)$ directly, NES maximizes the expected value of the loss function under the search distribution. This allows for gradient estimation in much fewer queries than typical finite-difference methods.

Andrew Ilyas choose a search distribution of random Gaussian noise around the current image $x$; that is, we have $\theta = x + \sigma\delta$, where $\delta\sim N(0,{I})$. we employ antithetic sampling to generate a population of $\delta_i$ values: instead of generating $n$ values $\delta_i \in N(0,{I})$, we sample Gaussian noise for $i \in \{ 1,...,\frac{n}{2}\}$ and set $\delta_j = -\delta_{n-j-i}$ for  $j \in \{ (\frac{n}{2}+1),...,n\}$. The following variance reduced gradient estimate:
\begin{equation}
    \nabla \mathbb{E} [ F(\theta)] \approx \frac{1}{\sigma n} \sum_{i=1}^n \delta_i F(\theta +\sigma\delta_i)
\end{equation}

Once the gradient has been estimated, the PGD can be applied directly to the attack:
\begin{equation}
    x^{(t)}=\prod_{[x_0-\epsilon,x_0+\epsilon]}(x^{(t-1)}-\eta \; sign(g_t))
\end{equation}
\end{spacing}
\textbf{Partial-information setting}
\\
\begin{spacing}{0.95}
In this setup, the attack does not start with image $x$, but with an instance $x_0$ of the target class $y_{adv}$, so that $y_{adv}$ will initially appear in the first $k$ classes. In each step $t$ of the update, there are 2 different steps. Ensure that the target category remains in the former top-k:
\begin{equation}
    \epsilon_t = \min \epsilon' s.t. rank (y_{adv}|\prod_{\epsilon'}(x^{(t-1)}))<k
\end{equation}
Maximize the probability value of the target class:
\begin{equation}
    x^{(t)}=\arg \mathop{max}\limits_{x'} P(y_{adv}|\prod{\epsilon_{t-1}}(x'))
\end{equation}
\end{spacing}
\begin{spacing}{1.3}
\textbf{Label-only setting}
\\
\end{spacing}
\begin{spacing}{0.97}
Consider a setting that assumes access to only the first $k$ sorted labels. As mentioned earlier, this case explicitly includes the setting $ k = 1$, but aims to design an algorithm that can include additional information when$ k > 1$.

The key idea behind the attack is that in the absence of an output score, an alternative way to characterise the success of the adversarial sample needs to be found. First, the authors define a discretized score $R(x(t))$ of the adversarial sample to quantify the degree of adversariality of the image at each step $t$ simply based on the ranking of the adversarial label $y_{adv}$:

\begin{equation}
    R(x^{(t)})= k-rank(y_{adv}|x^{(t)})
\end{equation}

As a proxy for the softmax probability, the robustness of the adversarial image to random perturbations is considered, using a discretization score to quantify the adversarial nature:

\begin{equation}
    S(x^{(t)}) = E_{\delta\sim u[-\mu,\mu]}[R(x^{(t)}+\delta)]
\end{equation}

A final Monte Carlo approximation was used to estimate this proxy score:

\begin{equation}
    \hat{S}(x^{(t)})= \frac{1}{n}\sum_{i=1}^n R(x^{(t)}+\mu \delta_i)
\end{equation}
\end{spacing}
Andrew Ilyas show the effectiveness of these algorithms by attacking an ImageNet classifier and demonstrate targeted adversarial examples for the Google Cloud Vision API, showing that our methods enable black-box attacks on real-world systems in challenging settings. Our results suggest that machine learning systems remain vulnerable even with limited queries and information.

\begin{spacing}{0.95}
\subsubsection{One pixel attack}

DNN-based methods surpass traditional image processing techniques, but recent research has found that adversarial images can allow classifiers to misclassify, by adding some perturbation to the image that is not recognizable to the human eye. Mainstream work adds perturbations that are full-image and easily recognized by the human eye, so Jiawei Su et al focus on proposing a single-pixel black-box attack method using differential evolution, where the only information available is the probabilistic label \cite{Su_2019}. The main advantages of this work compared to previous work are:
\end{spacing}
Effectiveness. On the Kaggle CIFAR-10 dataset, non-targeted attacks could be launched by modifying just one pixel on three common deep neural network structures, with success rates of 68.71\%, 71.66\% and 63.53\%. We also found that each natural image could be scrambled to the other categories of 1.8, 2.1 and 1.5. On the original CIFAR-10 dataset, where the attack scenarios are more limited, we show success rates of 22.60\%, 35.20\% and 31.40\%. On the ImageNet dataset, non-targeted attacks on the BVLC AlexNet model were also possible by changing a single pixel, showing that 16.04\% of the test images could be attacked.

Semi-black box attack. Only black-box feedback is required, without internal information about the target DNN, such as gradients and network structure. Our approach is also simpler than existing methods because it does not abstract the problem of searching for perturbations into any explicit objective function, but focuses directly on increasing the probabilistic label values of the target class.

Flexibility.Enables attacks on networks where gradients are difficult to calculate and not microscopic.
\begin{spacing}{0.95}
Generating an adversarial image can be formulated as an optimization problem with constraints. Assume that the input image can be represented by a vector, where each scalar element represents a pixel. Let $f$ be a target image classifier receiving an n-dimensional input and $x = (x_1,...,x_n)$ be the original natural image correctly classified into class $t$. Thus, the probability that $x$ belongs to class $t$ is $f_t(x)$. The vector $e(x) = (e_1,...,e_n)$ is an additional added adversarial perturbation with a target class $adv$ and a maximum alteration range $L$ . Note that $L$ is generally measured by its length. So the objective of the target adversarial attack is to find an optimised solution $e(x)^*$ : 

\begin{align}
    \text{maximize}_{e(x)^*}& \  f_{adv}(x+e(x)) \\
    \text{s.t.}& \ \Vert e(x)\Vert \le L \notag
\end{align}
\begin{spacing}{0.95}
The problem involves finding two values: (a) which dimensions need to be perturbed and (b) the range of modifications for each dimension. Therefore, the above equation has some modifications:
\end{spacing}
\begin{align}
    \text{maximize}_{e(x)^*}& \  f_{adv}(x+e(x)) \\
    \text{s.t.}& \ \Vert e(x)\Vert_0 \le d \notag
\end{align}
\end{spacing}
Whereas the usual adversarial image is constructed by perturbing all pixels while imposing an overall constraint on the cumulative modification intensity, the minority pixel attack considered by Jiawei Su is the opposite, focusing on only a few pixels but not limiting the modification intensity.

They encode the perturbations into a tuple (candidate solution) which is optimised (evolved) by means of a difference operation. A candidate solution contains a fixed number of perturbations and each perturbation is a tuple that contains five elements: the $x-y$ coordinates and the RGB value of that perturbation. A perturbation modifies one pixel. The initial number of candidate solutions (children) is 400 and in each iteration another 400 candidate solutions (children) will be generated using the conventional DE formula:

\begin{equation}
    x_i(g+1) = x_{r_1}(g)+F(x_{r_2}(g)-x_{r_3}(g)), \ r_1\neq r_2\neq r_3
\end{equation}

where $x_1$ is an element of the candidate solution, $r_1$, $r_2$, $r_3$ are random numbers, $F$ is a scale parameter set to 0.5 and $g$ is the index of the current generation. Once generated, each candidate solution will compete with its corresponding parent based on the overall index, with the winner surviving to the next iteration. 

\subsection{Transfer-based attack}
\subsubsection{Momentum is added to the iterative process} \label{transfer-opt-start}

In order to increase the migration ability of adversarial samples, domestic and foreign scholars have proposed a variety of methods that combine optimization-based adversarial attacks with various means. 

The first approach is to add momentum to the iteration. Yinpeng Dong \cite{dong2018boosting} proposed the use of gradient momentum to enhance the iterative attack of FGSM, namely MI-FGSM. Similar to the effect of gradient descent, momentum can further improve the success rate of black box attacks by accumulating velocity vectors in the gradient direction of the loss function, thereby stabilizing the direction of renewal of the perturbation and helping to get rid of poor local extreme values to produce more examples of transferable confrontation. This momentum method leads to faster convergence and less oscillation.
\begin{spacing}{0.95}
MI-FGSM can be written as:
\begin{equation}
    \begin{aligned}
        &g_{i+1}=\mu g_i+\frac{\nabla_xJ\left(x_i^A,y\right)}{\|\nabla_x J\left(x_i^A,y\right)\| _1}\\
        &x_{i+1}^A=x_i^A+\alpha sign\left(g_{i+1}\right)
    \end{aligned}
\end{equation}
Where $g_{i+1}$ represents the accumulated gradient momentum after the ith iteration, $\mu $ is the attenuation factor of the momentum term. Since the gradient obtained in multiple iterations is not of the same magnitude, the current gradient $\nabla_{\rm x}J\left(x_i^A,y\right)$ obtained in each iteration is normalized.\cite{dong2018boosting}

In the black box model, most existing attacks can’t generate strong adversarial samples to against the defense model. Jiadong Lin (2020) proposed a new idea. He regarded the generation process of adversarial samples as an optimization process, and proposed a new method to improve the mobility of adversarial samples --Nesterov iterative fast gradient sigh method (NI-FGSM).

NI-FGSM is a fusion of NAG and I-FGSM. Nesterov Accelerated Gradient (NAG), a variation of the common gradient descent method, can accelerate the training process and significantly improve the convergence.

NAG can be written as:
\begin{equation}
    \begin{aligned}
        &{\rm v}_{t+1}=\mu {\rm v}_t+\nabla_{\theta _t}J\left(\theta_t-\alpha \mu {\rm v}_t \right)\\
        &\theta _{t+1}=\theta _t-\alpha {\rm v}_{t+1}
    \end{aligned}
\end{equation}
Where ${\rm v}_{t+1}$ and ${\rm v}_t$ are the update directions of this time and last time respectively. $\nabla_{\theta _t}J\left(\theta_t-\alpha \mu {\rm v}_t \right)$ represents the gradient of the objective function at $\left(\theta_t-\alpha \mu {\rm v}_t \right)$. The hyperparameter $\mu $ is the attenuation weight of the last update direction, so it is generally between 0 and 1, and $\alpha $ is the learning rate. This formula indicates that the direction of each parameter update not only depends on the gradient of the current position, but also is affected by the direction of the last parameter update.
\end{spacing}
Compared with MI-FGSM, in addition to stabilizing the updating direction, NAG makes use of the second derivative information of the target function, which helps the attack to effectively look ahead. This property helps us to escape from poor local optimal solutions more easily and quickly, thus improving transferability.

NI-FGSM can be written as:
\begin{equation}
    \begin{aligned}
        &x_{t}^{nes}=x_{t}^{adv}+\alpha \mu g_t\\
        &g_{t+1}=\mu g_t+\frac{\nabla_xJ\left(x_t^{nes},y^{true} \right)}{{\|\nabla_x J\left(x_t^{nes},y^{true} \right)\| _1}}\\
        &x_{t+1}^{adv}=Clip_X^\epsilon \left\{x_t^{adv}+\alpha sign\left(g_{t+1}\right) \right\}
    \end{aligned}
\end{equation}
where $g_t$ denotes the accumulated gradients at the iteration t, and $\mu $ denotes the decay factor of $g_t$.

It should be noted that before each iteration, NI-FGSM accumulates the gradient in the previous direction and then updates it.

\subsubsection{The input transformation is considered in the iterative process}

The second method is to input the transformation during the iteration. Spatial transformation includes rotation, enlargement and reduction, which is a kind of data expansion method. The purpose is to enrich the representation of similar data by using spatial transformation of input images. This can not only prevent the adversarial examples from overfitting the model, but also improve the portability of the adversarial examples.

The study shows that the defense model's resistance to transferable adversarial examples is mainly due to the fact that the defense model has different identification areas to predict compared with the normal training model. While normal models have similar attention maps, defense models produce different attention maps(defense models are either trained under different data distributions or transform inputs before classification). When FGSM, BIM or other models are used to generate adversarial samples, only a single sample is optimized, so it will be highly correlated with the gradient of the identification area or the white box model at the attack point of the input data. For another black box model with different identification areas, the adversarial sample is difficult to maintain adversarial.

In order to reduce the impact of different recognition regions among models, Yinpeng Dong \cite{dong2019evading} proposed a transition invariant attack method (TIM). In this paper, a collection of images and their translated images are generated adversarial sample. It is hoped that the adversarial sample are insensitive to the recognition area of the attacked white box model and have a higher probability of deceiving another black box model with a defense mechanism. However, to generate such a sample, the gradient of all images in the set needs to be calculated, which is a large amount of computation. In order to improve efficiency, it is proposed in the literature that, under certain assumptions, the convolution gradient method is made for untransferred images, in which the convolution kernel is predetermined. This method can be combined with any gradient-based attack method (such as FGSM, etc.) to generate more transferable and adversarial samples.

TIM can be written as 
\begin{equation}
    \begin{aligned}
        &{\rm {TI-FGSM}}:x^{adv}=x^{real}+\epsilon sign\left(W\ast\nabla_x J\left(x^{real},y \right) \right)\\
        &{\rm {TI-BIM}}:x_{t+1}^{adv}=x_t^{adv}+\alpha sign\left(W\ast\nabla_x J\left(x_t^{adv},y \right) \right)
    \end{aligned}
\end{equation}
Where W is the Gaussian convolution kernel.\cite{dong2019evading}

In addition, Cihang Xie \cite{xie2019improving} proposed DIM, a diverse Input method. He adopted the idea of data enhancement. Before inputting images into the model, random transformation of input samples was carried out, such as random adjustment of sample size or random filling of given distribution. Then the converted image is input to the classifier, and the subsequent gradient calculation is carried out to generate the adversarial sample. This method can be combined with MI-FGSM, etc., which can also improve the migration of attacks.

Optimization has another way of extending the model. It is found that DNN may have scale invariance in addition to translation invariance. Specifically, the loss values of the original and scaled images on the same model are similar. Therefore, scaling can be used as a model extension method.
\begin{spacing}{0.95}
SIM is based on the discovery of scale-invariant properties of deep learning models. Jiadong Lin \cite{lin2020nesterov} proposed to optimize adversarial perturbation by scaling input images to avoid "overfitting" the white box model under attack and generate more adversarial examples that could be transferred.
\end{spacing}
\subsubsection{Train the auxiliary classifier}\label{transfer-opt-end}
The third approach is to train an auxiliary classifier. Almost all current adversarial attacks of CNN classifiers rely on information derived from the output layer of the network, Nathan Inkawhich\cite{inkawhich2020transferable} proposed a new adversarial attack based on the modeling and exploitation of class-wise and layer-wise deep feature distributions, which called the Feature Distribution Attack (FDA).

The basic idea is that, to compute FDA adversarial noise from layer l, we first build a composite model using the truncated white-box model $f_l$ and the corresponding layer’s auxiliary model $g_{l,c\cdot }=y_{tgt\cdot }$($g_{l,c\cdot }$ can capture the layer-wise and class-wise feature distributions, aiming to model the probability that the layer l features extracted from input x are from the class c feature distribution. )The loss is calculated as the Binary Cross Entropy (BCELoss) between the predicted $p\left(y=y_{tgt}| f_l\left(x\right)\right)$ and 1. Thus, we perturb the input image in the direction that will minimize the loss, in turn maximizing $p\left(y=y_{tgt}| f_l\left(x\right)\right)$, to generate targeted (or un-targeted) adversarial examples. The key intuition for the targeted methods is that if a sample has features consistent with the feature distribution of class c at some layer of intermediate feature space, then it will likely be classified as class c. 

FDA can be written as:
\begin{equation}
    \max \limits_{\delta} p\left(y=y_{tgt}| f_l\left(x+\delta \right)\right)
\end{equation}
where $\delta $ is a perturbation of the “clean” input image x.

The paper stress that unlike standard attacks that use output layer information to directly cross decision boundaries of the white-box, the FDA objective leverages intermediate feature distributions which do not implicitly describe these exact boundaries.\cite{inkawhich2020transferable}
\begin{spacing}{0.95}
Nathan Inkawhich \cite{inkawhich2020transferable} proposed to significantly improve the FDA method by extending it into a more flexible framework to allow for multilayer perturbations across the intermediate feature space, including the output layer. 
\end{spacing}
The difference is that each $g_{l,y_{tgt}}$is a feature distribution model that estimates the probability that the layer l feature map $f_l\left(x\right)$ is from a sample of class $y_{tgt},i.e.\quad p\left(y_{tgt}| f_l\left(c\right)\right)$. For a chosen $y_{tgt}$ and set of layers, we accumulate the losses $w.r.t.\quad y_{tgt}$ at each intermediate layer and the output layer. By optimizing the sum, we are noising x with $\delta $ such that $x+\delta $ lies in high probability regions of the target class at several layers across feature space. We find that this method significantly improves transferability of the generated adversarial samples.
\begin{spacing}{0.95}
This method is written as:
FDA can be written as:
\begin{equation}
    \max \limits_{\delta} p\left(y=y_{tgt}| f_l\left(x+\delta \right)\right)+\eta \frac{\| f_l\left(x+\delta \right)-f_l\left(x\right)\| _2}{\| f_l\left(x\right)\| _2}-\gamma H\left(f\left(x+\delta \right),y_{tgt}\right)
\end{equation}
The second component is the feature disruption term, which enforces that the feature map of the perturbed input is significantly different than the feature map of the original input.

A key contribution in this work is the third component to include multi-layer information. One way to do so is to incorporate the white-box model’s output prediction as part of the attack objective. Let $H\left(f\left(x\right),y\right)$ be the standard cross-entropy loss between the predicted probability distribution f(x) and the target distribution y (commonly a one-hot distribution). We include this term in the FDA objective to create FDA+xent.
\end{spacing}
where $\gamma >0$, weighting the contribution of cross-entropy term. This attack objective optimizes the noise such that the layer l feature map is in a high-probability region of the target class, and that the output prediction of the white-box model is of the target class. Addition of the cross-entropy term is because of the probability distribution over the classes, as measured in the “optimal transfer layer,” has low correlation with the probability distribution over the classes at the output layer. 

The paper introduced a feature space-based adversarial attack framework that allows for perturbations along the extracted feature hierarchy of a DNN image classifier to achieve state-of-the-art targeted blackbox attack transferability. These performance gains are attributed to the inclusion of multi-layer information, which leads to significantly higher disruption in the feature spaces of both the white-box and black-box. \cite{DBLP111111}

Other methods include transforming the target function and then using optimization algorithms to search for solutions to the target function.

\subsubsection{Generative-based adversarial transfer}\label{transfer-gen}

Studies have shown that perturbations exist in large contiguous regions rather than scattered in multiple small discontinuous pockets, so the most important thing when generating perturbations is to consider the direction of perturbations, rather than specific points in space\cite{goodfellow2014explaining}. Considering generative perturbation modeling for a given region of classes, Mopuri et al.\cite{mopuri2018nag} make the GAN framework a suitable choice for our task by introducing a generative model similar to GAN to capture the distribution of unknown adversarial perturbations as well as the space of unknown perturbations, which is not affected by parametric assumptions, and the target distribution is unknown (there are no known samples in the target distribution of adversarial perturbations). This method successfully trains a generator network to capture unknown target distributions without the need for any training samples. The resulting model produces adversarial perturbations with a large diversity for migration attacks almost immediately, and the method can effectively simulate perturbations that spoof multiple deep models at the same time.

For perturbations that need to be generated, two types of adversarial perturbations can be considered: generic and instance-dependent. The perturbations that a particular instance depends on can vary depending on the image in the datasets. To generate these perturbations, we need a function to acquire a natural image and output a hostile image, and we can approximate this function with a deep neural network. A general perturbation is a fixed perturbation that, when added to a natural image, significantly reduces the accuracy of the pre-screened network. For the generation of adversarial perturbations for specific instances such as image dependencies, four types of adversarial attacks are generally considered: target general, non-target generic, target image dependency, and non-target image dependency. While the non-target black-box transferability of adversarial perturbations has previously been extensively studied, changing the decision of an invisible model to a specific "target" category remains a challenging feat. Under the condition of small perturbation norm, high deception rates have been achieved for all tasks, and these perturbations can be successfully transferred between different target models. A method\cite{poursaeed2018generative} is proposed to discard pre-trained models and only use generators to generate adversarial examples to accommodate perturbations of input samples, which avoids the need for iterative gradient computation and allows us to quickly generate perturbations, in addition to using generative models to create adversarial perturbations, allowing us to further train more complex models. Moreover, the study also proves that the resulting perturbations can be transferred in different models, which is a migration attack.

\subsubsection{Universal adversarial perturbations}\label{uap}

\textbf{Universal adversarial perturbations}
\\
\begin{spacing}{0.95}
Traditional optimized counter attacks typically target a single sample, while the goal of a generic counter attack is to find a generic counter disturbance that works for as many samples as possible \cite{mopuri2017fast}. That is, finding a general perturbation $v$ such that for most images $x$,
\end{spacing}
\begin{equation}
\hat{k}(x+v) \ne \hat{k}(x) \text{for ``most" } x\sim\mu
\end{equation}
There are also constraints to ensure visual quality and attack effectiveness:
\begin{equation}
\|v\|_p\leq\xi
\end{equation}
\begin{equation}
{P}_{x\sim\mu}(\hat{k}(x+v)\neq\hat{k}(x))\geq1-\delta
\end{equation}
Seyed-Mohsen et al \cite{moosavi2017universal} propose a systematic algorithm for computing universal perturbations, and show that state-of-the-art deep neural networks are highly vulnerable to such perturbations, although the human eye cannot distinguish. This article generates general perturbations through Deep fool . The purpose of this algorithm is to find the minimum perturbation v, so that xi+v moves out of the correct classification region Ri. This algorithm makes it easy to calculate the corresponding disturbances for different models such as VGG, Google LeNet, ResNet, and so on. The article demonstrates the existence of these general perturbations (as shown in the figure below) and their good generalization.

\begin{spacing}{0.95}
The implementation of this algorithm is: given the dataset $X=\{x_1,..., x_m\}$. Based on constraints and optimization objectives, the algorithm iterates over data points on dataset $X$ to gradually construct a generic perturbation. Iterative solution is to calculate a new perturbation based on the previous perturbation $v$. The specific optimization formula is as follows:
\begin{equation}
    \triangle  v_{i}\gets {\arg\min}_{r}
    \|r\|_2\\ s.t \hat{k}(x_i+v+r) \ne \hat{x_i}
\end{equation}

\begin{figure}[h]
    \caption{gradient transformer module}
    \centering
    \includegraphics[width=4cm]{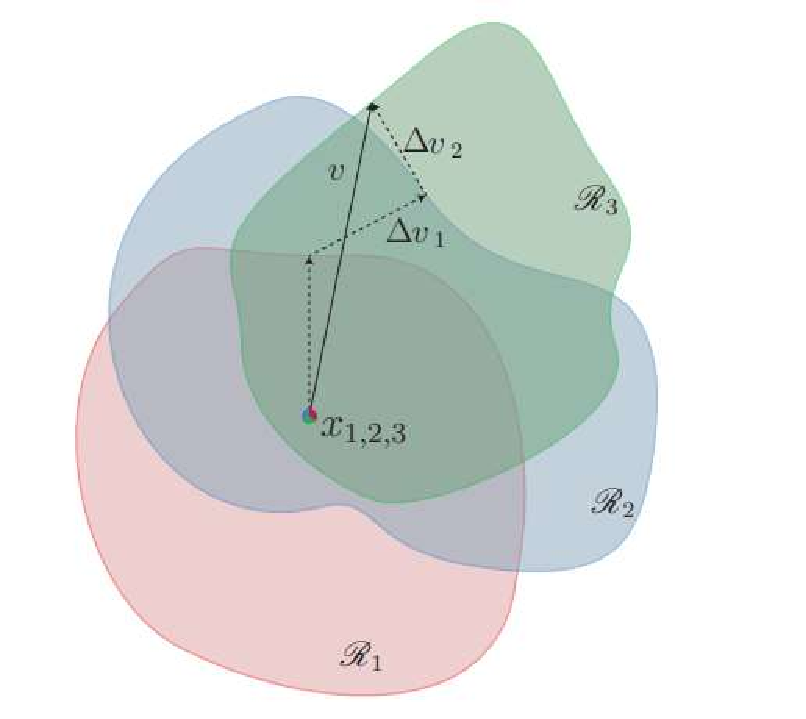}
    \label{fig:} 
\end{figure}
\vspace*{1cm}

To ensure that constraints are met $\|v\|_p\leq\xi$.The updated general perturbation is further projected to a radius of $\xi$, centered at 0 on the ball $\ell_p$. Therefore, the projection operation is defined as
\begin{equation}
    P_{p,\xi}(v)= {\arg\min}_{v'}
    \|v-v'\|_2\\ s.t \|v\|_p\leq\xi
\end{equation}
The actual update process is $v\gets P_{p,\xi}(v+\triangle  v_{i}) $
\\
\end{spacing}
\textbf{Fast feature fool}
\\

 Konda Reddy Mopuri \cite{mopuri2017fast} provides a method for effectively generating generic perturbations by deceiving features learned in multiple network layers without data, thereby misleading CNN.
 And these disturbances have good generalization. Fast feature fool is used to generate generic perturbations without relying on data, which fool CNN by supersaturating the features of multi-layer learning (replacing the "flip the label" target), that is, by adding perturbations to the input, destroying the features of each layer to mislead the features of subsequent layers. Cumulative disturbances along the network hierarchy will make the network unable to distinguish between original inputs, resulting in a large number of prediction errors at the final layer.
 Its essence is to find a disturbance that can generate the maximum false activation at each layer without any data provided by the target cnn.The objective function of this optimization problem is:
 \begin{equation}
     Loss=-log(\prod_{i=1}^{K} \overline{l_{i}}(\delta)) 
     \quad s.t \quad \|\delta\|_{\inf}<\xi
 \end{equation}

where $l_i(\delta)$ is the mean activation in the output tensor at layer $i$ when $\delta$ is input to the CNN. Note that the activations are considered after the non-linearity (typically ReLU),therefore it is non-negative. K is the total number of layers in the CNN at which we maximize activations for the perturbation $\delta$.$\xi$ is the limit on the pixel intensity of the perturbation $\delta$.

\begin{spacing}{0.95}
Considering all convolutional layers before the fully connected layer, the product of multi-layer average activation is calculated to maximize all layers simultaneously. To avoid the occurrence of extreme values, the $log()$ function is applied to the product. Because optimization updates are perturbations, Therefore, there is no need for sample data to participate in the process of generating generic countermeasures. After each update, the disturbance is clipped to the constraint range to meet the constraint of disturbance size.
The perturbation proposed by researchers has three characteristics. One is that the same perturbation can deceive multiple images based on a target dataset on a given CNN; The second is to prove the transferability of disturbances between multiple networks trained on the same dataset; Third, they surprisingly retain the ability to deceive cellular neural networks trained on different target datasets (compared to relying on data). Experiments have shown that data independent universal adversarial disturbances may pose a more serious threat than data dependent disturbances.\\
\end{spacing}

\begin{spacing}{1.2}
\textbf{Generalizable data-free objective for crafting universal adversarial perturbations}
\\
\end{spacing}
\begin{spacing}{0.99}
GD-UAP is also a general data-independent anti-perturbation meth-od proposed by Mopuri et al \cite{mopuri2018generalizable}, which is an extension of the Fast Feature Fool method. the difference with Fast Feature Fool method is that it uses the i-th layer tensor activation value $L_2$ norm instead of the average activation value,and the loss function is :
\end{spacing}
\begin{equation}
     Loss=-log(\prod_{i=1}^{K} \|l_{i}(\delta)\|_{2}) 
     \quad s.t \quad \|\delta\|_{\inf}<\xi
\end{equation}

 The author uses this method in target detection, semantic segmentation and other tasks to achieve a variety of attacks. 
 This paper proves that the disturbance is data-free. The author conducted a comprehensive analysis of the proposed goals, including: a thorough comparison of the GD-UAP approach with the opposing portions of the dependent data, and an assessment of the strength of UAPs in the presence of various defense mechanisms. It extended UAP attacks to visual tasks other than image classification and proposed a target model for training data distribution with minimal prior information to create stronger disturbances.\\

\textbf{Regionally homogeneous perturbations}
\\
\begin{spacing}{0.95}
Yingwei Li et al \cite{li2020regional} proposed Regionally Homogeneous Perturbations using a natural training model and a defense model for white box attacks.In this paper, a gradient transformer module is proposed to obtain regional homogeneity confrontation samples. The principle is to increase the correlation of pixels in the same region, thereby causing regional uniform disturbance.

Based on this observation, researchers proposed a transformation paradigm and a gradient transformer module to generate area uniform perturbation (RHP) specifically for attack defense.Researchers have demonstrated the effectiveness of regional homogeneous disturbances through experiments attacking a series of defense models. Using semantic segmentation tasks to attack and test object detection tasks has proven the cross task portability of RHP.

\section{Adversarial Attack in Computer Vision Applications}\label{obj-dec}
\subsection{Adversarial attack in object detection}

Object detection \cite{2021A1111} is one of the research hotspots in the field of computer vision, which aims at detecting and classifying targets, specifically in the given specific environment, for example, a static image or a dynamic video data set, in different background in the image or video of interest in the specific location of the object of interest in the extraction of information and with a rectangular box marked, and the class of the object is given.

\begin{figure}[htbp]
    \centering
    \includegraphics[scale=0.45]{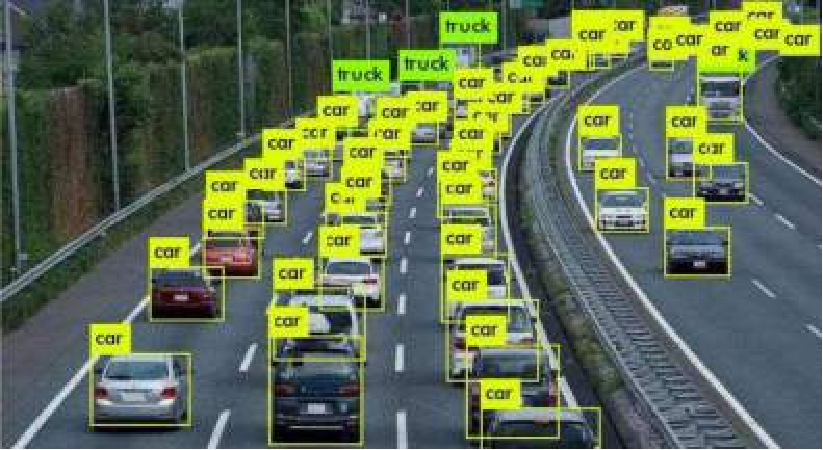}
    \caption{Object detection is used in vehicle detection}
    \label{obj}
\end{figure}

Figure \ref{obj} shows an example of using target detection technology to detect vehicles on the road.
\vspace*{0.7cm}

Despite the rapid development of target detection, recent studies have shown that target detection has shortcomings in security \cite{2021A} and is easily deceived by antagonistic samples. 
\end{spacing}
\subsubsection{Two-stage network attack}

In 2017, Lu et al \cite{2017Adversarial} proposed the DFooL method to mislead the corresponding detector by adding disturbances to the ``Stop” sign and the face image, which is the first article in the field of target detection to adversarial sample generation. In the same year, Xie \cite{2017Aaadversarial} and others proposed Dense Adversary Generation (DAG) , which attacks the target candidate set generated on the RPN component of two-stage network, each target candidate region is assigned an antagonism tag and a gradient rise strategy is executed until the candidate region is successfully predicted to stop iteration for the specified antagonism tag. DAG is one of the most classical attack methods in target detection. In the actual attack, the effect of DAG is better, but it is time-consuming because of the need of iterative attack on each candidate box. Shapeshifter \cite{2018Robus11111t} is the first targeted attack method against Faster R-CNN proposed by Chen et al., which borrows from the adversarial attack Methods CW and Expectation Over Transformation (EOT) \cite{2018Synthesizing} in image classification, the adversarial sample generated by the stop sign attack successfully deceived Faster R-CNN, but the attack required modification of the entire stop sign at a high cost. Li et al put forward a RAP attack on two-stage network \cite{2018Robust}. In the attack, by designing a loss function which combines the loss of classification and the loss of location, compared with DAG method, Li's method makes use of the location information of target detection to attack and the attack method is original, but the actual attack new ability is more general and the attack mobility against RPN is poor. Zhang et al (Cap) \cite{2020Contextual}, on the basis of the previous attack methods, make full use of the context information, extract the context information of the target object in the image and destroy these areas, at the same time, the background area score is improved to increase the intensity of the attack, and better experimental results are obtained on PASCAL VOC and MS COCO datasets.

\subsubsection{One-stage network attack}

On one-stage network model, Song et al (PhyAttack) \cite{2018Physical} inspired by the Robust Physical Perturbations (RP2) in the image classification domain \cite{2017Robust} added an additional antagonistic loss function to the original RP2, the YOLOv2 detector was successfully deceived by gradient descent to reduce the mark score in the image so that the detector could not detect the stop sign. Wang (Daedalus) \cite{2021Daedalus} attacks on the common module Non Maximum Suppression (NMS) of the one-stage detection model, which makes YOLO network produce false alarm, false negative and so on. Liao (CA) et al \cite{2020Category} were the first to attack a particular anchor-free class model CenterNe in a single-stage network by looking for important pixel areas in an image, it uses its high-level semantic information to carry on the category attack to the detector. The CenterNet model is used to extract thermograms in which the pixel points with a score greater than the threshold are defined as the important pixel areas and their prediction categories. During the attack, these important pixel points are attacked to deviate from the original category, the resulting adversarial samples can not only attack the anchor-free model, but also successfully migrate to the traditional one-stage and two-stage target detection models.

\subsubsection{Both detectors can be attacked}

Wei \cite{2018Transferable} proposed a UEA (Unified and Efficient Adversary) method to solve the problems of high time-cost and poor migration of target detection adversarial samples, combining GAN with the advanced classification loss and the low-level feature loss, the GAN is trained to generate the adversarial samples. The experiment proves that the adversarial samples can be generated in real-time by the trained GAN network, and the real-time attack on video is realized, the generated adversarial samples have a high attack success rate against SSD and YOLO in single-phase networks. Wu et al \cite{2019G} improved and migrated the UAP (Universal Adversarial perturbation) \cite{2017Universal} in image classification to the field of target detection and proposed the G-UAP model.By selecting a batch of pictures to attack the data set and attacking each picture at the same time as well as by reducing the confidence score of foreground objects in the picture and increasing the confidence score of background, the corresponding disturbance can be obtained, finally, all the perturbations are aggregated as the feature map of the network, so that we can learn the general perturbations from this batch of images to deceive more images.Chow et al \cite{2020TOG} proposed an iterative TOG method that can attack both two-stage and one-stage target detectors simultaneously, according to the final attack effect, TOG method can be divided into three categories: Target Vanishing, forged tag, and classification error, in the process of iteration generation the original image of gradient modification is obtained by performing gradient descent on the set loss function until the attack is successful or the iteration number is reached. TOG can achieve nearly one hundred percent attack success rate on white box, but it has poor mobility.

In addition to these global adversarial-jamming attacks, some other researchers have proposed a new adversarial, which only adds perturbations in one region of the original image so that the perturbations in that region can affect the whole image which can described as the local perturbation attack to achieve the goal of spoofing the target detector.

\subsubsection{Local adversarial-jamming}

In 2018, Li et al \cite{2018Attacking} first proposed the method BPatch for local adversarial-jamming attacks on two-stage detectors. In this method, the target detector is attacked by adding disturbing blocks to the background outside the target of the image. BPatch is also an attack against RPN (area proposal network) which is a unique part of the two-phase detector. Since the RPN network generates a large number of candidate fields containing the candidate boxes, the next stage of the network will arrange the candidate boxes for the RPN network according to the confidence level, the candidate boxes above the confidence threshold were selected for the next stage of classification and location regression.BPatch presents an attack method for RPN network filtering mechanism by reducing the confidence of the high confidence candidate obtained by the RPN layer and make the candidate boxes that end up in the next layer of the network contain little or no foreground targets. Liu et al \cite{2018DPatch} add a patch to the picture and treat the patch as a GT (Ground Truth) check box.Back propagation causes the network to optimize the patch directly so that the final detector is affected by the patch, resulting in a detection error. In addition, Wang et al. \cite{2020An} proposed a particle swarm optimization target detection black box attack EA which uses a natural optimization algorithm to guide disturbance generation in place, but this approach is time-consuming. Thys et al (Adversarial-yolo) \cite{2019Fooling} have created an adversarial block to fool the YOLOv2 detector so that the YOLOv2 detector can not detect the presence of a person after the person has carried the adversarial block.

\subsection{Adversarial attacks in semantic segmentation}\label{seg}

As an extension of classification, adversarial attacks are also widely used in the field of semantic segmentation. Semantic segmentation is a fundamental task in computer vision that refers to the use of deep learning models to segment different objects in an image from the pixel level and label each pixel with a category. For example, we may need to distinguish all pixels belonging to cars in an image and paint them in a specific color.Fully Convolutional Networks (FCN)\cite{1} is a framework for semantic segmentation of images proposed by Jonathan Long et al. in 2015, which is the pioneering work of deep learning for semantic segmentation. In this paper, full convolution is used instead of full concatenation, and the operation of inverse convolution is used to recover information in order to compensate for the loss caused by the downsampling operation, and high and low channel features are fused to recover low-level visual information. The state-of-the-art semantic segmentation models are usually based on standard image classification architectures\cite{2}, extended by additional components such as dilated convolution\cite{3}, specialized pooling\cite{4}, skip connections\cite{1}, conditional random fields (CRFs)\cite{5} and/or multiscale processing\cite{6}, but their impact on robustness has never been thoroughly investigated.
\par Arnab et al\cite{7} applied an adversarial attack method migration for image classifiers to semantic segmentation models, carefully evaluating work on adversarial attacks on semantic segmentation models such as DeepLab V2\cite{6} and PSPNet\cite{4}, showing that segmentation models built on deep neural networks are also plagued by adversarial samples. And Xie et al\cite{2017Aaadversarial} proposed DAG (Dense Adversary Generation) attack to start applying adversarial attacks to target detection and semantic segmentation, and proposed an adversarial attack method with migration capability. DAG is a greedy algorithm that considers all targets simultaneously and optimizes the overall loss function by simply specifying for each target a adversarial label and iteratively performs gradient backpropagation to obtain cumulative perturbations, but does not minimize the number of considered paradigms. To address this, Cisse et al. proposed the Houdini attack\cite{10} for several tasks including semantic segmentation. The goal of this approach is to maximize the surrogate loss for a given perturbation budget (i.e., a constraint on the L$\infty$-norm), so that no minimum perturbation is generated, and by directly formulating adversarial samples for the combined infeasible task loss, it can be achieved to deceive any gradient-based learning machine.Ozbulak et al. studied adversarial examples in medical image segmentation tasks\cite{11}. They proposed a targeted attack, which is equivalent to a conventional penalty approach. However, the weight of the penalty term is set to 1, which leads to a large L2 parametrization, even compared to DAG. \cite{2019Fooling} generated an "adversarial patch" by optimizing the target loss, this small patch can cause the target to evade the AI detector of the object. \cite{13} proposed a scenario-specific attack that uses the CARLA driving simulator to improve the transferability of EOT-based attacks in real 3D environments. Instead of the traditional cross-entropy loss, \cite{14} designed a loss function that can attack image regions far from the patch, which contains several separate loss terms that do not contain the patch pixels, with the aim of gradually shifting the deception focus from increasing the number of misclassified pixels to increasing the antagonistic strength of the patch on the misclassified pixels to improve the attacker's ability to induce pixel misclassification., and the paper also validates the effectiveness of the scenario-specific attack. \cite{15} proposed a segmentation attack method called "segPGD", and the experimental results showed that the convergence was faster and better than that of PGD.
\par In addition there are more studies on the robustness of semantic segmentation under adversarial attacks. Since previous work on deep neural networks performing vulnerably against samples has focused on classification tasks and has rarely been used on large-scale datasets and structured tasks such as semantic segmentation, based on this, \cite{17} generated adversarial samples by invoking the last-likely method proposed by Kurakin et al. and misclassified each pixel in the pixel to the closest class for a more natural effect in order to analyze how adversarial perturbations affect semantic segmentation. \cite{18} generate generic perturbations by "letting the network produce a fixed target segmentation as the output" and "keeping the segmentation constant except for removing the specified target class" to make the network produce the desired target segmentation as the output, which can change the semantic segmentation of an image in a near-arbitrary The perturbation can change the semantic segmentation of the image in a nearly arbitrary way. It is important to know that the generic perturbation was first proposed in \cite{2017Universal}, which is a fixed perturbation with a specific algorithm that iterates over a sample of the target set, while \cite{20} creates generic attacks and image-related attacks by using an end-to-end generative model instead of an iterative algorithm, with a significant improvement in generation and inference time. \cite{21} then compared the advantages and disadvantages of local perturbation with generic perturbation, where is a perturbation generated indirectly by a noise function and an intermediate variable to make the gradient of pixels propagate infinitely. \cite{22} observed that spatial consistency information can potentially be exploited to robustly detect adversarial examples even when a strongly adaptive attacker has access to the model and the detection strategy, and that the adversarial examples of the attacks considered in the paper hardly migrate between models.
\par Much of the above work uses FGSM\cite{23} (Fast Gradient Sign Attack) or its derivative models, which provide a coarse robustness evaluation and are not a minimization attack. \cite{24} proposed a white-box attack which is based on approximate partitioning to produce adversarial perturbations with smaller L1, L2, or L$\infty$-norm. This attack can handle a large number of constraints within a non-convex minimization framework by augmenting Lagrangian methods and adaptive constraint scaling and masking strategies.
\par For certain adversarial attacks designed for classification problems, especially those that do not rely on projecting to the estimated decision boundary (e.g., DeepFool or FAB\cite{26}), they can also be applied to the segmentation domain, such as PGD, DDN\cite{27}, FMN\cite{28}, PDGD and PDPGD\cite{29}, and ALMA\cite{30}, where PDPGD\cite{29} although it relies on approximate partitioning, it uses the AdaProx algorithm\cite{31}. adaProx introduces a mismatch between the scale and the step size of the gradient step in the computation of the approximation operator, and the convergence of this algorithm in the non-convex case is well worth investigating.

\end{document}